\documentclass[lettersize,journal]{IEEEtran}

\usepackage[T1]{fontenc}
\usepackage{graphicx}
\usepackage{pdfpages}
\usepackage{amsmath,amssymb,amsfonts}
\usepackage{booktabs}
\usepackage{multirow}
\usepackage{array}
\usepackage{float}
\usepackage{stfloats}%
\usepackage{textcomp}
\usepackage{url}
\usepackage{cite}%
\usepackage[dvipsnames]{xcolor}
\usepackage{xspace}
\usepackage{enumitem}
\usepackage{microtype}

\definecolor{linkblue}{rgb}{0,0.3,0.6}
\usepackage[breaklinks,hidelinks]{hyperref}
\hypersetup{colorlinks=true,citecolor=linkblue,linkcolor=linkblue,urlcolor=linkblue}

\newcommand{\etal}{\textit{et al}.\xspace}

\makeatletter
\renewcommand{\paragraph}{\@startsection{paragraph}{4}{0ex}%
  {1.6ex plus 0.6ex minus .2ex}{-0.8em}{\normalfont\normalsize\bfseries}}
\makeatother
\newcommand{\pp}{\,pp}
\newcommand{\ci}[2]{{\scriptsize[#1,\,#2]}}
\newcommand{\gd}[1]{\textbf{#1}}
\definecolor{dropc}{HTML}{8C510A}%
\definecolor{gainc}{HTML}{55A868}
\newcommand{\drop}[1]{\textcolor{dropc}{#1}}
\newcommand{\gain}[1]{\textcolor{gainc}{#1}}

\newcommand{\repodurum}{will be released at}

\newcommand{\repo}{\url{https://github.com/7unahan/privacy_treshold}}

\makeatletter
\long\def\@makecaption#1#2{%
\ifx\@captype\@IEEEtablestring%
\footnotesize\bgroup\par\centering\@IEEEtabletopskipstrut%
{\normalfont\footnotesize #1}\\{\normalfont\footnotesize #2}\par
\addvspace{0.5\baselineskip}\egroup\@IEEEtablecaptionsepspace
\else%
\@IEEEfigurecaptionsepspace
\setbox\@tempboxa\hbox{\normalfont\footnotesize {#1.}\nobreakspace #2}%
\ifdim \wd\@tempboxa >\hsize%
\normalfont\footnotesize {#1.}\nobreakspace #2\par%
\else%
\hbox to\hsize{\normalfont\footnotesize\hfil\box\@tempboxa\hfil}%
\fi\fi}
\makeatother

\begin{document}

\title{Wrong Input, Wrong Threshold:\\Re-evaluating Adversarial Facial Privacy}

\author{Tunahan Parlay\i c\i,~Erchan Aptoula,
        and~Y\"ucel Sayg\i n%
\thanks{T.~Parlay\i c\i\ was supported by the Scientific and Technological
Research Council of T\"urkiye (T\"UB\.ITAK) under the 2244 Industrial PhD
Fellowship Program, project 119C045.}%
\thanks{The authors are with the Faculty of Engineering and Natural Sciences,
Sabanc\i\ University, 34956 Istanbul, T\"urkiye
(e-mail: \nolinkurl{tunahanp@sabanciuniv.edu};
\nolinkurl{erchan.aptoula@sabanciuniv.edu};
\nolinkurl{ysaygin@sabanciuniv.edu}). \emph{Corresponding author: Tunahan
Parlay\i c\i.}}%
\thanks{The measurement code, the per-pipeline threshold calibration and the
survey data of Tab.~\ref{tab:survey} \repodurum\ \repo.}}%

\markboth{IEEE TRANSACTIONS ON INFORMATION FORENSICS AND SECURITY, VOL.~XX, NO.~XX, MONTH~2027}%
{Parlay\i c\i\ \MakeLowercase{\textit{et al.}}: Wrong Input, Wrong Threshold: Re-evaluating Adversarial Facial Privacy}

\maketitle

\begin{abstract}
Published face-privacy methods report protection above $90\%$ against
black-box recognition models. That number comes from an evaluation protocol
with two independent defects, reported in a measure that cannot size either
one. \emph{The input is wrong}: methods published since $2022$ feed the
network the whole photograph, or a detector's box, where deployed systems
detect, align to a landmark template and crop. Re-scoring the same files
through a realistic front end costs a standard transfer attack $56\%$ of its
dodging protection and a budget-allocating attack $7\%$, with no image
regenerated. \emph{The measure is wrong}: that loss is a difference of success
rates, so it saturates, moving by $8.2\times$ over a budget sweep while the
effect beneath it moves by $1.34\times$. We therefore report the paired cosine
shift, in which both thresholds cancel algebraically, so the input defect
holds without committing to a calibration. Measured that way it is
$2.2\times$ and $10.7\times$ larger on three unseen recognizers, and at
matched budget it reverses the $\ell_\infty$ transfer ranking: the one
surveyed method that aligns at evaluation gains $0.18$ where MI-FGSM loses
$0.05$. The cause is a \emph{resampling mismatch}, not perturbation lost
outside the crop; optimising through a differentiable alignment operator
largely removes it, adding $+24$ to $+58$ points on unseen recognizers and a
commercial system, with a residual on one of them. \emph{The thresholds are wrong}: the four constants
inherited since $2021$ hold the false-accept rate near its nominal $1\%$ on
aligned crops but admit $2.3$--$5.4\%$ on whole images, with an
\emph{opposite-signed} bias across dodging and impersonation, the two tasks
reported side by side. Differentiable alignment and gallery scoring are not
new: the first generation of this literature used both, and we show what
dropping them cost.
\end{abstract}


\section{Introduction}
\label{sec:intro}

\begin{figure}[t]
  \centering
  \includegraphics[width=\linewidth]{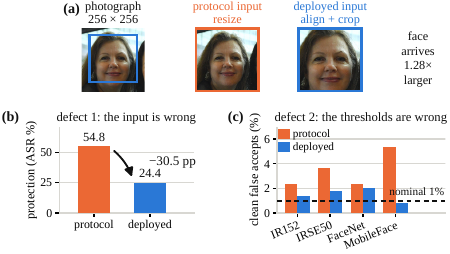}
  \caption{\textbf{Two defects in one protocol.} \textbf{(a)}~The
  $112{\times}112$ tensor a recognizer receives from one photograph under the
  \emph{evaluation protocol} of published work (whole-image resize) and under a
  \emph{deployed} pipeline (detect, align, crop; footprint outlined).
  \textbf{(b)}~The same files moved to the deployed pipeline (MI-FGSM, FFHQ,
  $N{=}1000$, thresholds per pipeline). \textbf{(c)}~The inherited thresholds
  admit $2.3$--$5.4\%$ clean false accepts against a nominal $1\%$.}
  \label{fig:teaser}
\end{figure}

People publish photographs of themselves, and face-recognition systems index
them without consent. A growing literature answers this with \emph{adversarial
facial privacy}: a small, ideally imperceptible perturbation added to the
photograph so that recognition models no longer match it to the person
\cite{shan2020fawkes,cherepanova2021lowkey,yang2021tipim,hu2022amtgan,sun2024diffam}.
The reported numbers are striking. Protection success runs above $90\%$ against
black-box models, measured by \emph{leave-one-out} transfer across four
standard recognition networks. These numbers are measured, not claimed, and we
do not dispute the measurements. We dispute what they measure, in two
independent ways, and we dispute the unit they are measured in.

\paragraph{The input is wrong}
Every method published since 2022 that we examined evaluates protection by
resizing the \emph{whole photograph}, or at most a detector's bounding box, to
the network's input resolution. No deployed pipeline we are aware of does this: deployed systems
run a detector, estimate five landmarks, apply a similarity transform to a
canonical template, and crop
\cite{deng2019arcface,zhang2016mtcnn,deng2020retinaface}. The perturbation is
therefore evaluated on a tensor the recognizer will never see. We push the same
protected images through the realistic front end, bit for bit and with no
re-optimisation. That costs a plain MI-FGSM transfer attack $56\%$ of its measured
protection ($30.5\pp$ of $54.8$) and a budget-allocating attack $7\%$
($6.9\pp$ of $92.9$; FFHQ \cite{karras2019stylegan}, $N{=}1000$, thresholds per
pipeline). On three modern recognizers no attack used as a surrogate the same
files lose $90$--$98\%$ and $20$--$69\%$, and across all three datasets the
transfer attack loses $46$--$98\%$ while the allocating attack's loss tracks
the dataset rather than the recognizer (Sec.~\ref{sec:modern}).

A reader could conclude that protection simply does not survive real pipelines.
It does. A telescoping decomposition (Sec.~\ref{sec:decomp}) attributes only
$-0.4\pp$ to cropping and $-9.5\pp$ to resampling geometry for the
budget-allocating attack. The aligned crop re-samples the face at $1.28\times$
the rate of the whole-image resize, so a perturbation tuned to one grid is
attenuated on the other. The failure is a \emph{loss of alignment, not of
information}, and optimising through a differentiable alignment operator
largely removes it (Sec.~\ref{sec:method}).

\paragraph{The measure is wrong}
Both figures above are differences of success rates, and a success rate is
bounded at both ends. Where protection is near $0\%$ under both front ends
there is nothing to lose, and near $100\%$ nothing to gain, so the difference
collapses towards zero in both regimes whatever the underlying effect. Sweeping
the perturbation budget makes this concrete: $|K_1|$ moves by a factor of $8.2$
and peaks in the middle of the range, where the effect beneath it moves by
$1.34$ (Sec.~\ref{sec:dcos}). We therefore report the paired shift in cosine
similarity, $\Delta\!\cos$, derived so that the two thresholds cancel
algebraically. It does not saturate, it needs no calibration, and it makes the input
defect independent of the threshold defect we turn to next: a reader who
rejects our calibration entirely still has to account for $\Delta\!\cos$.

\paragraph{The thresholds are wrong}
A threshold is a property of a score distribution, which depends on the front
end. The four constants this field has inherited since 2021 are
reused across methods with mutually incompatible front ends. They hold the
false-accept rate near its nominal $1\%$ on aligned crops but admit
$2.3$--$5.4\%$ on whole images (Sec.~\ref{sec:threshold}). The error is
opposite-signed, because dodging counts similarity below the threshold and
impersonation above it, so one too-low threshold makes the protocol look worse
for one task and better for the other. Correcting it doubles our measured
protocol drop and more than halves two published impersonation results. The two
defects reach the two tasks unequally, which we state here and not in the
limitations: the input defect is established on dodging, where at calibrated
thresholds impersonation shows no separable drop, while impersonation is
carried by the threshold defect. This literature tabulates its headline numbers
on impersonation, so both defects are needed to account for what it reports.

\paragraph{Two mechanisms, not one}
Shifting the sampling grid by half an output pixel, at identical scale and
crop, costs MI-FGSM up to $55.5\pp$ and the budget-allocating attack at most
$3.5\pp$ (Sec.~\ref{sec:mechanism}). High-frequency attacks live in the
aliasing phase of one resampling grid; low-frequency ones are shift-robust and
lose margin only under scale mismatch. The same tests expose \emph{grid
leakage}: direct evaluation inflates apparent transfer whenever surrogate and
target share a resampling grid. One remedy addresses both regimes.

This is a paper about evaluation first and method second: the largest
correction ($+50.2\pp$) is on plain MI-FGSM, and the mechanism is base-attack
agnostic by construction.

\paragraph{Contributions}
\begin{enumerate}\itemsep1pt
\item \textbf{The protocol misranks, it does not only inflate.} The one
surveyed method that aligns at evaluation time is understated by $25.2\pp$,
more than the spread separating the other three, so the published ordering
does not survive the correction. The survey behind that claim reads the
evaluation pipelines of thirteen published methods (2021--2026) from released
code at a stated commit: they differ from each other, no method published
after $2021$ whose evaluation we could read aligns, and nine copy one threshold
table. Three of the thirteen release no evaluation script, and we say so in the
table rather than scoring them as if we had read one.
\item A like-for-like measurement of what survives a realistic front end
(three datasets, two detectors, two conventions, four transfer attacks, three
unseen modern recognizers and a gallery of independent photographs of each
subject), with a decomposition of the loss.
\item \textbf{A threshold-free reading of the protocol gap.} Subtracting two
cosines is not a new statistic. What is new is that the gap can be stated
without the calibration this paper disputes: the field's unit, a difference of
success rates, saturates at both ends of its range; the paired
cosine shift of Eq.~\ref{eq:dcos} does not, and the two thresholds cancel in it
algebraically, so the input defect can be established without committing to any
calibration. Measured that way the gap is nearly budget-invariant ($1.34\times$) where the
success rate difference varies by $8.2\times$, and it is $2.2\times$ and
$10.7\times$ larger on recognizers no attack used as a surrogate.
\item \textbf{The inherited thresholds do not fit the pipeline they are applied
to}, with
the induced error opposite-signed across the two tasks; and \textbf{two loss
mechanisms}, separated by controlled experiments, with the grid-leakage effect
that inflates direct evaluation.
\item Pipeline-aware optimisation: a drop-in change to the embedding path that
removes most of the loss on two attack families and is worth $+24$ to $+58$ points of
deployed protection on a commercial system and on three unseen modern
recognizers. A second commercial endpoint reproduces both, though its scores
admit no single $1\%$-FAR operating point. The measurement code, including the
calibration and the front ends, \repodurum\ \repo.

\end{enumerate}

\section{Related Work}
\label{sec:related}

\paragraph{Adversarial facial privacy}
Fawkes \cite{shan2020fawkes} and LowKey \cite{cherepanova2021lowkey} introduced
perturbing one's own photographs before posting them, and TIP-IM
\cite{yang2021tipim} formulated it as targeted identity protection under a
natural-image constraint. AMT-GAN \cite{hu2022amtgan} carries the perturbation
in transferred makeup and introduces the threshold table later work reuses;
CLIP2Protect \cite{shamshad2023clip2protect} searches a generative latent space
under text guidance, DiffAM \cite{sun2024diffam} replaces the generator with a
diffusion model, and the line continues through 2024--2026 with diffusion,
text-guided and pose-invariant variants
\cite{advcpg2025,liu2024advdiffusion,li2024gift,li2025tca,salar2025weakening,le2025diffprivate,wang2026protego}.
We compare evaluation pipelines rather than protection strength
(Tab.~\ref{tab:survey}) and re-measure two of these methods end-to-end
(Sec.~\ref{sec:published}).

\paragraph{What the first generation already did}
LowKey \cite{cherepanova2021lowkey} optimises \emph{through} a differentiable
detection-and-alignment operator, and scores protection as rank-1 and rank-50
retrieval against a gallery of \emph{other} photographs of the same identity,
with no verification threshold; TIP-IM aligns at evaluation
(Tab.~\ref{tab:survey}). Both practices then leave the literature. Neither the
operator nor the gallery protocol is ours to claim; what we add is the
diagnosis of what dropping them costs, and through which mechanism.

\paragraph{Evaluation as the object of study}
Adversarial machine learning has repeatedly found a published number measuring
something other than it claimed: obfuscated gradients made weak defences look
robust \cite{athalye2018obfuscated}, adaptive attacks broke defences their own
evaluations had passed \cite{tramer2020adaptive}, and a community checklist
exists for the same reason \cite{carlini2019evaluating}. Those critiques
concern the \emph{attack} an evaluation runs; ours the \emph{input} it runs on
and the \emph{threshold} it scores with, both upstream of any attack.

\paragraph{Robustness of adversarial examples to preprocessing}
Adversarial examples have been studied under input transformations since the
first gradient attacks \cite{goodfellow2015fgsm,madry2018pgd}: JPEG
\cite{guo2018countering,zhang2023lfap,guo2024jpegrobust}, resizing and padding
\cite{xie2018mitigating}, with EOT the standard remedy \cite{athalye2018eot}.
Convolutional networks are themselves unexpectedly sensitive to sub-pixel
translation \cite{azulay2019generalize,zhang2019antialias}, the clean-image
counterpart of the grid effect we measure on perturbations. Closest to us,
Croft \etal\ \cite{croft2025detectors} evaluate adversarial face images across
seven detectors and find the detector choice matters. They crop the detected
region and leave alignment out of scope, which preserves the inter-detector
variation that five-point alignment largely \emph{absorbs}
(Sec.~\ref{sec:unseen}); their perturbations are optimised on that crop, so a
shifted detection is damage where for a whole-image perturbation the same shift
helps (Sec.~\ref{sec:decomp}). Their finding that the interpolation kernel is
immaterial at fixed geometry agrees with our account: what matters is which
region is sampled at which scale. Their remedy, randomised crops and rescaling,
is of the kind of our scale-only control, which recovers half the available
gain (supplement).

\paragraph{Thresholds as inherited constants}
Verification thresholds are reported as calibrated operating points
(FAR@$0.01$), but here they are passed from paper to paper as constants
(Tab.~\ref{tab:survey}), and to our knowledge no prior work reports the
false-accept rate they actually produce on the pipeline they are applied to;
Sec.~\ref{sec:threshold} does.

\paragraph{Dodging versus impersonation}
Zhou \etal\ \cite{zhou2024rethinking} show that in the black-box setting a
successful impersonation does not guarantee a successful dodge, so impersonation
overstates privacy. We add a measurement reason: at the inherited thresholds the
two tasks are biased in opposite directions (Sec.~\ref{sec:threshold}).

\paragraph{Relation to our concurrent work}
The budget-allocating attack we evaluate \cite{parlayici2026ssmattack} is our
own earlier work, under review elsewhere at the time of writing. It is used
unmodified and is no contribution of
this paper: it is reported there under the protocol we show to be defective,
and our correction costs it $6.9\pp$ rather than crediting it.

\paragraph{Recognition pipelines}
Margin-based losses \cite{deng2019arcface,wang2018cosface} and efficient
backbones \cite{chen2018mobilefacenets} define the attacked models; MTCNN
\cite{zhang2016mtcnn} and RetinaFace \cite{deng2020retinaface} with the
five-point ArcFace template are the de facto front end, which makes the gap a
property of the protocol rather than of one system.

\section{The Wrong Input}
\label{sec:gap}

\subsection{What published work actually does}
\label{sec:survey}

\begin{table}[tb]
\caption{\textbf{Evaluation front ends, read from released code.} Thirteen
methods over six years: the front ends disagree, no method published after
$2021$ \emph{whose evaluation we could read} aligns, and ten score themselves
against the \emph{same} four constants,
whose earliest datable appearance is
\cite{yin2021advmakeup} and which reappear to six decimals in seven later code
bases. No paper in the chain says where the values came from.
\emph{ev.}\ is the evidence each row rests on: \textsf{C}~evaluation code read,
\textsf{A}~no evaluation script released, so the row reports the attack code's
front end, \textsf{P}~no code released, so the row comes from the paper.
$^\ast$ stated in the paper, absent from the code; $^\ddagger$ the only
aligning method, and the only one the protocol \emph{understates}
($K_1 = \gain{+25.2\pp}$); $^\S$ scores a pre-aligned dataset whole, so it
never detects or aligns itself. The supplement gives the per-method evidence
with file, line and commit.}
\label{tab:survey}
\centering\small
\setlength{\tabcolsep}{3.2pt}
\resizebox{\columnwidth}{!}{%
\begin{tabular}{@{}llccc@{}}
\toprule
Method & Year & Front end at eval. & Thr. & ev. \\
\midrule
TIP-IM \cite{yang2021tipim}$^\ddagger$      & '21 & detect $+$ \gd{align} & own & C \\
Adv-Makeup \cite{yin2021advmakeup}           & '21 & whole image$^\S$      & \gd{earliest} & C \\
AMT-GAN \cite{hu2022amtgan}                  & '22 & bounding-box crop     & copied & C \\
CLIP2Protect \cite{shamshad2023clip2protect} & '23 & bounding-box crop     & copied & C \\
DiffAM \cite{sun2024diffam}                  & '24 & whole image           & copied & C \\
Adv-Diffusion \cite{liu2024advdiffusion}     & '24 & whole image           & copied & C \\
GIFT \cite{li2024gift}                       & '24 & whole image           & copied$^\ast$ & A \\
Adv-CPG \cite{advcpg2025}                    & '25 & not documented        & copied$^\ast$ & P \\
TCA$^2$ \cite{li2025tca}                     & '25 & whole image           & copied & A \\
Salar \etal\ \cite{salar2025weakening}       & '25 & bounding-box crop     & copied & C \\
DiffPrivate \cite{le2025diffprivate}         & '25 & whole image           & own & C \\
DiffProtect \cite{liu2026diffprotect}        & '26 & whole image           & copied & C \\
Protego \cite{wang2026protego}               & '26 & bounding-box crop     & none & C \\
\bottomrule
\end{tabular}}

\end{table}

We read the released evaluation code of thirteen methods (Tab.~\ref{tab:survey}).
Three facts stand out. (i) The front ends are mutually inconsistent, so these
papers' headline numbers are not commensurable even before the question of
realism arises. (ii) Exactly one method aligns, and it is the earliest; the
best any later one does is crop a detector's bounding box. (iii) Ten
nonetheless score themselves against the \emph{same} four thresholds, and not
as a historical artefact: that table is published in $2021$ and still copied
verbatim into code in $2025$. Two published tables make the consequence visible without
any experiment of ours: Adv-CPG \cite{advcpg2025} and DiffAM
\cite{sun2024diffam} score the \emph{same} method on the \emph{same} dataset at
the \emph{same} nominal thresholds, yet differ by up to $23.7\pp$ on a single
model (supplement). The protocol is not a fixed object.

\subsection{Protocol and notation}
\label{sec:setup}

Let $x$ be a photograph and $\delta$ the protective perturbation. We write
$P_{\mathrm{dir}}(x)$ for the \emph{protocol} front end, a bilinear resize of
the whole image to the network input, and $P_{\mathrm{ali}}(x)$ for the
\emph{deployed} one, which detects five landmarks, fits the similarity
transform $M$ to the ArcFace template by the Umeyama estimator
\cite{umeyama1991}, then warps and crops. Protection is measured in the
standard leave-one-out black-box setting: optimised on three of \{IR152,
IRSE50, FaceNet, MobileFace\}, evaluated on the fourth, reported as the mean
over the four held-out targets. Dodging success is the fraction of protected
photographs whose cosine similarity to the subject's own clean photograph falls
\emph{below} the threshold. The clean reference is embedded through the
\emph{same} front end, since a deployed system enrols through its own pipeline.
Where the detector returns no face the protection counts as successful: a
system that cannot find the face cannot match it.

We use FFHQ \cite{karras2019stylegan}, CelebA-HQ \cite{karras2018pggan} and
LADN \cite{gu2019ladn}, with the sample size of each measurement given in
Sec.~\ref{sec:exp}. On the two that need resizing we audit our own
preprocessing the way we ask others to: the recovered perturbation's sup-norm
reads $16.00/255$ against a nominal $16/255$ under the interpolation the
generating code used and $53/255$ under a different one, so the choice is
checkable rather than assumed (supplement). Confidence intervals are bootstrap $95\%$ over
\emph{images} \cite{efron1979bootstrap}; since $K_1$ is a difference of two
proportions on the same images, we resample the paired difference rather than
the two arms independently. Per-target significance is an exact McNemar test
\cite{mcnemar1947}, computed \emph{per target} rather than on the four-target
mean, which can hide a target where the effect is absent.

\paragraph{Normalisation conventions}
Every result is reported in two input conventions: the \emph{corrected} one
gives each network its own published preprocessing, the \emph{field} one what
the released code does (supplement). The choice moves individual targets by up
to $25\pp$ but largely cancels in the four-target mean, and no conclusion here
depends on it.

\subsection{A threshold-free measure of the gap}
\label{sec:dcos}

$K_1$ is a difference of two success rates, so it is bounded by the floor and
the ceiling of the rate itself: near $0\%$ under both front ends there is
nothing to lose, near $100\%$ nothing to gain. Our own budget sweep makes this
concrete, $|K_1|$ varying by a factor of $8.2$ across $\epsilon\in[8,32]/255$
and peaking in the middle (Tab.~\ref{tab:eps}): such a quantity cannot say how
large the gap is, only whether it is visible at one operating point.

We therefore add a measure that does not saturate, derived so that the
thresholds cancel.
Write the \emph{margin} of a protected photograph under front end $P$ as
$m_P = \tau_P - \cos(f_P(x{+}\delta), f_P(x))$, positive exactly when dodging
succeeds, so that the paired shift is
$\Delta m = (\tau_{\mathrm{ali}} - \tau_{\mathrm{dir}}) -
(\cos_{\mathrm{ali}} - \cos_{\mathrm{dir}})$. An \emph{unprotected}
photograph has $\delta = 0$, both cosines equal one, and $\Delta m$ reduces to
$\tau_{\mathrm{ali}} - \tau_{\mathrm{dir}}$: a property of the two
calibrations alone, carrying nothing about the protection, so we subtract it.
What remains is
\begin{equation}
\Delta\!\cos \;\equiv\; \Delta m - (\tau_{\mathrm{ali}} - \tau_{\mathrm{dir}})
\;=\; \cos_{\mathrm{dir}} - \cos_{\mathrm{ali}} ,
\label{eq:dcos}
\end{equation}
in which no threshold appears. $\Delta\!\cos < 0$ means the protected
photograph stays \emph{more} similar to its own identity under the deployed
front end, that is, the protocol overstates protection; $\Delta\!\cos > 0$ means
the protocol understates it.

Three properties make this the measure we report first. It is
threshold-free, so no conclusion about the input defect depends on the
calibration we criticise in Sec.~\ref{sec:threshold}; we verified
Eq.~\ref{eq:dcos} numerically on all $35$ stored runs, where the two sides
agree to machine precision. It does not saturate. A difference of cosines is bounded in
principle, by $[-2,2]$, but nothing measured here comes near that range,
whereas a difference of rates is compressed whenever either rate sits at its
floor or its ceiling, which is where these methods operate. And it is commensurable across front ends, the impostor
standard deviation changing by at most $10\%$ between them on all seven
recognizers, and by at most $4\%$ on six of the seven (supplement). We keep $K_1$ throughout, because the literature
reports success rates, and we mark the cells where it sits at the floor or the
ceiling.

$\Delta\!\cos$ sizes the \emph{protocol gap}, the change a front end makes to a
score. It is not an operational privacy metric: a shift of a given size does
not convert to a verification or identification error without a threshold and
a population, and the same shift can matter more on one recognizer than on
another. The operational claims therefore rest on the calibrated rates of
Sec.~\ref{sec:threshold}, on the identity-labelled calibration repeated there,
and on the gallery of independent photographs of each subject
(Tab.~\ref{tab:gallery}); $\Delta\!\cos$ is what makes those claims comparable
across front ends.

A deployed system detects on the photograph it receives, so the aligned arm
takes its landmarks from the \emph{protected} image; the clean-image
convention makes the gap \emph{larger} by up to $0.022$, in the same direction
in $29$ of $30$ measurements and by $0.001$ in the exception, for the reason
given in Sec.~\ref{sec:decomp}. We report the smaller one (supplement has
both).

\begin{table}[tb]
\caption{\textbf{How the gap varies, and what the success rate hides.}
\emph{(a)} Four targets, $50$ iterations, thresholds per front end. $|K_1|$
varies by $8.2\times$ and the relative loss by $15.2\times$; $|\Delta\!\cos|$
by $1.34\times$ (unrounded). The ends of the $K_1$ row are floor ($3.2\%$) and ceiling
($96.3\%$) effects. TI-FGSM behaves like MI-FGSM ($1.58\times$); DI-FGSM
randomises scale during optimisation, a weak form of the remedy of
Sec.~\ref{sec:method}, and has no gap to lose at any budget
($|\Delta\!\cos|\leq0.011$; supplement).
\emph{(b)} ``mism.''\ is the resampling mismatch of Sec.~\ref{sec:decomp};
``3 mod.''\ excludes MobileFace (see text). The FFHQ row is the main set of
Tab.~\ref{tab:k1} ($N{=}1000$), the other two $N{=}150$. The panels are
separately generated runs and are read within, not across. No threshold
enters $\Delta\!\cos$ in either (Eq.~\ref{eq:dcos}).}
\label{tab:eps}
\centering\small
\setlength{\tabcolsep}{3.6pt}
\resizebox{\columnwidth}{!}{%
\begin{tabular}{@{}lcccccc@{}}
\multicolumn{7}{@{}l}{\emph{(a) along the budget}, MI-FGSM, FFHQ, $N{=}300$}\\
\toprule
$\epsilon$ ($/255$) & 8 & 12 & 16 & 21 & 26 & 32 \\
\midrule
$P_{\mathrm{dir}}$ success & 3.2 & 25.2 & 59.6 & 84.5 & 93.2 & 96.3 \\
$K_1$ (\pp)                & \drop{$-2.7$} & \drop{$-11.0$} & \drop{$-22.2$}
                           & \drop{$-20.1$} & \drop{$-10.7$} & \drop{$-5.2$} \\
relative loss (\%)         & 82.1 & 43.5 & 37.3 & 23.8 & 11.6 & 5.4 \\
$\Delta\!\cos$             & \gd{$-0.030$} & \gd{$-0.037$} & \gd{$-0.037$}
                           & \gd{$-0.034$} & \gd{$-0.032$} & \gd{$-0.028$} \\
\bottomrule
\end{tabular}}

\vspace{3pt}
\resizebox{\columnwidth}{!}{%
\begin{tabular}{@{}lccc|ccc@{}}
\multicolumn{7}{@{}l}{\emph{(b) along the dataset}, $\Delta\!\cos$, stored sets}\\
\toprule
& \multicolumn{3}{c|}{MI-FGSM $16/255$} & \multicolumn{3}{c}{allocation, peak $26$} \\
\cmidrule(lr){2-4}\cmidrule(lr){5-7}
Dataset & mism. & 4 mod. & 3 mod. & mism. & 4 mod. & 3 mod. \\
\midrule
FFHQ      & 1.28 & $-0.109$ & \gd{$-0.134$} & 1.28 & $-0.013$ & \gd{$-0.063$} \\
CelebA-HQ & 1.31 & $-0.086$ & \gd{$-0.126$} & 1.31 & $-0.002$ & \gd{$-0.075$} \\
LADN      & 1.10 & $-0.080$ & \gd{$-0.112$} & 1.10 & $+0.053$ & \gd{$-0.011$} \\
\bottomrule
\end{tabular}}

\end{table}

\subsection{Measuring the gap}
\label{sec:measure}

\begin{table*}[!tb]
\caption{\textbf{$K_1$ and $K_2$ in one view.} Identical protected images
scored through both front ends; nothing is regenerated. ASR \%, mean over four
leave-one-out targets, thresholds calibrated at FAR$=1\%$ \emph{per pipeline}
and dataset, bootstrap $95\%$ CI on the paired difference; pipeline-aware
optimisation removes the drop on both datasets and conventions
($p\leq10^{-11}$ per target). $^\S N{=}300$. $^\ddagger$Separate run, own
calibration ($K_1$ $-28.0$, $+6.3$); do not difference against $P_{\mathrm{dir}}$.}
\label{tab:k1}
\centering\small
\setlength{\tabcolsep}{4pt}
\begin{tabular}{@{}lcccc|ccc|c@{}}
\toprule
& \multicolumn{4}{c|}{FFHQ, $N{=}1000$, corrected convention}
& \multicolumn{3}{c|}{CelebA-HQ, $N{=}150$} & FFHQ, field conv. \\
\cmidrule(lr){2-5}\cmidrule(lr){6-8}\cmidrule(lr){9-9}
Protection & $P_{\mathrm{dir}}$ & mtcnn & retina & $K_1$
           & $P_{\mathrm{dir}}$ & $P_{\mathrm{ali}}$ & $K_1$ & $K_1$ \\
\midrule
MI-FGSM ($\epsilon{=}16/255$) & 54.8 & 24.4 & 25.2$^\ddagger$ & \drop{$-30.5$} \ci{-32.2}{-28.8} & 72.3 & 34.3 & \drop{$-38.0$} \ci{-42.2}{-33.8} & \drop{$-25.6$} \ci{-27.1}{-24.1} \\
\;$+$ \gd{ours}               & 68.2 & 74.6 & 73.0$^\ddagger$ & \gain{$+6.4$} \ci{+4.9}{+7.8} & 85.7 & 86.7 & $+1.0$ \ci{-1.8}{+3.8} & \gain{$+9.3$} \ci{+7.8}{+10.7} \\
\midrule
uniform budget$^\S$           & 89.7 & 72.5 & --   & \drop{$-17.2$} \ci{-20.5}{-14.0} & -- & -- & -- & -- \\
Allocation \cite{parlayici2026ssmattack}    & 92.9 & 86.0 & 84.5 & \drop{$-6.9$} \ci{-8.0}{-5.8} & 97.8 & 89.3 & \drop{$-8.5$} \ci{-11.0}{-6.2} & \drop{$-2.1$} \ci{-3.3}{-0.9} \\
\;$+$ scale EOT               & 91.9 & 92.2 & 91.6 & $+0.3$ \ci{-0.6}{+1.2} & -- & -- & -- & -- \\
\;$+$ \gd{ours}               & 95.9 & 98.2 & 98.1 & \gain{$+2.4$} \ci{+1.7}{+3.0} & 98.5 & 99.2 & $+0.7$ \ci{-0.5}{+2.0} & \gain{$+8.9$} \ci{+7.9}{+9.9} \\
\bottomrule
\end{tabular}

\end{table*}

Tab.~\ref{tab:k1} applies both front ends to the same protected images. Every
existing protection drops significantly, and the ordering is informative: the
weaker and spatially flatter the perturbation, the more it loses. Plain MI-FGSM
loses $56\%$ of its measured protection, the uniform-budget variant $19\%$, and
the allocation-based attack $7\%$ (MI-FGSM's figure falls to $36\%$ when
regenerated with each surrogate's own preprocessing; supplement). These shares
are read at one budget and do not transfer to another, since the relative loss
is itself budget-dependent (Tab.~\ref{tab:eps}); the ordering, not the
percentages, is what we rely on, and Sec.~\ref{sec:when} measures it on an
axis that separates the two mechanisms.

\begin{table}[tb]
\caption{\textbf{Telescoping decomposition} (FFHQ, $N{=}300$). Each column adds
one element of the deployed front end and reports the marginal change in ASR.
The dominant term is resampling, not cropping; rotation is close to inert;
landmark displacement \emph{helps}. Thresholds calibrated at FAR$=1\%$ per
configuration; bootstrap $95\%$ CI on the total. The totals reproduce the $K_1$
of Tab.~\ref{tab:k1} ($-6.9$, $-17.2$, $-30.5$), which the
decomposition never sees and which for two rows is a different sample
($N{=}1000$).}
\label{tab:decomp}
\centering\small
\setlength{\tabcolsep}{3.4pt}
\resizebox{\columnwidth}{!}{%
\begin{tabular}{@{}lccccc@{}}
\toprule
Protection & crop & $+$resample & $+$rotation & $+$lm.\ shift & total \\
\midrule
Allocation            & $-0.4$ & \drop{$\mathbf{-9.5}$}  & $-0.4$ & \gain{$+3.8$} & $-6.6$ \ci{-8.5}{-4.7} \\
uniform budget        & $-3.6$ & \drop{$\mathbf{-17.8}$} & $-0.3$ & \gain{$+4.4$} & $-17.3$ \ci{-19.8}{-15.1} \\
MI-FGSM               & $-4.7$ & \drop{$\mathbf{-26.6}$} & $-1.5$ & \gain{$+1.8$} & $-30.9$ \ci{-33.8}{-28.1} \\
\bottomrule
\end{tabular}}

\end{table}

\subsection{Why: a causal decomposition}
\label{sec:decomp}

We insert the elements of the deployed front end one at a time, each
configuration differing from the previous in exactly one respect
(Tab.~\ref{tab:decomp}). The totals, $-6.6$, $-17.3$ and $-30.9\pp$, reproduce the
$K_1$ of Tab.~\ref{tab:k1} on a sample the decomposition never sees. The terms
are marginal contributions in this order, not unique causal effects; cropping is
nested in the aligned geometry, so we measure it on the whole-image one, and
Sec.~\ref{sec:mechanism} tests the resampling term independently. Each
configuration is scored at its own FAR$=1\%$ threshold, so a term carries the
transform together with the recalibration it forces; that is the comparison we
want, at matched false-accept rate, but it is a reason to read the terms as an
ordering rather than as isolated magnitudes.

\textbf{(i) It is not the crop.} The aligned crop covers $60.7\%$ of the
photograph, yet restricting the attack to that footprint is nearly free for the
allocation-based attack ($0.4\pp$): once an allocator has concentrated the
budget on the face, discarding $39\%$ of the image area discards almost no
useful perturbation. Its squared perturbation energy inside the footprint is $1.60\times$ the area
share, against $1.03\times$ for MI-FGSM.

\textbf{(ii) It is the resampling.} Switching the resampling from the
whole-image grid to the aligned one costs $9.5$ to $26.6\pp$. The aligned crop
maps a $\sim$$199\,\mathrm{px}$ region of the $256\,\mathrm{px}$ photograph
onto $112\,\mathrm{px}$, where the protocol maps the full $256$: the face
arrives $1.28\times$ larger and its perturbation is sampled at a rate it was
not optimised for (Fig.~\ref{fig:teaser}). Sec.~\ref{sec:mechanism} shows this
account holds for the allocation-based attack and \emph{not} for MI-FGSM.

\textbf{(iii) Rotation is close to inert} ($\leq 1.5\pp$), which removes the
most visually salient difference between the pipelines from the explanation,
while \textbf{landmark displacement helps} ($+1.8$ to $+4.4\pp$): the
protection moves the detector's estimates by $2.56\,\mathrm{px}$ and the
resulting misalignment degrades the recognizer. The sign of that term is set by
where the perturbation was optimised, which is why Croft \etal\
\cite{croft2025detectors} find the opposite one on perturbations optimised on
the detected crop.

\subsection{When does the gap appear?}
\label{sec:when}

If the loss is a scale mismatch, a dataset framed like the aligned crop should
show no gap, and an attack whose loss is a grid effect should be indifferent to
the framing. Both predictions can be read off the same table once the measure
does not saturate (Tab.~\ref{tab:eps}b).

\textbf{The allocation-based attack follows the mismatch.} Its gap is
$-0.063$ on FFHQ and $-0.075$ on CelebA-HQ, whose mismatches are $1.28$ and
$1.31$, and $-0.011$ on LADN, whose mismatch is $1.10$: a factor of six
between the matched and the mismatched framing, with non-overlapping intervals.
This is the scale mechanism of Sec.~\ref{sec:mechanism} measured on three
datasets rather than inferred from one controlled sweep.

\textbf{MI-FGSM does not.} Its gap is $-0.134$, $-0.126$ and $-0.112$ on the
same three datasets, a spread of $1.2\times$ against the allocation attack's
$6\times$. A perturbation locked to one sampling grid loses that lock whenever
the grid changes, and the framing of the dataset does not change it. The
simpler reading, that MI-FGSM drops on every dataset, is correct
but weaker than what is visible here: the two attacks are not differently
sized versions of one effect, they are in different regimes, and the dataset
axis separates them.

\textbf{MobileFace is the exception, and it is systematic.}
On all three datasets and both attacks its $\Delta\!\cos$ has the opposite
sign to the other three recognizers ($-0.034$, $+0.035$, $+0.017$ for MI-FGSM
and $+0.136$, $+0.217$, $+0.244$ for the allocation attack), which is why
Tab.~\ref{tab:eps}b reports the pooled value with and without it. We do not treat this as noise.
MobileFace is also the recognizer whose two calibrations disagree most
($-0.102$ against $-0.016$ to $-0.033$; Tab.~\ref{tab:thr}): its whole-image
impostor distribution sits far higher than its aligned one, by $4.4\times$ the
largest shift of the other three, which leaves an attack less room under that
front end. Standardising each pipeline's cosine by its own impostor
distribution removes $51\%$ of the anomaly on the rows of this table, against
a ceiling of about $49\%$ in a simulation where a location shift generates it
entirely (supplement). That account covers these rows and not the
pipeline-aware variants, where the same test removes $15\%$ and we offer no
explanation. Three datasets
support an ordering and not a quantitative law, and the ordering is what we
claim.

\section{The Wrong Threshold}
\label{sec:threshold}

\begin{table}[tb]
\caption{\textbf{The inherited thresholds are compatible with the aligned
pipeline and not with the whole-image one.} The
four inherited constants (earliest in \cite{yin2021advmakeup}), beside thresholds calibrated to
FAR@$0.01$ \emph{per pipeline} on $500$ clean FFHQ images, pooling all
$124{,}750$ pairs, and the false-accept rate they produce against a
nominal $1\%$. Field convention; the corrected one gives the same structure
($2.4$--$3.8\%$ against $0.8$--$2.0\%$).}
\label{tab:thr}
\centering\small
\resizebox{\columnwidth}{!}{%
\begin{tabular}{@{}lrrrrr@{}}
\toprule
 & \multicolumn{3}{c}{threshold} & \multicolumn{2}{c}{clean FAR at inherited} \\
\cmidrule(lr){2-4}\cmidrule(lr){5-6}
Model & inherited & $P_{\mathrm{ali}}$ & $P_{\mathrm{dir}}$ & $P_{\mathrm{dir}}$ & $P_{\mathrm{ali}}$ \\
\midrule
IR152      & 0.167 & 0.179 & 0.202 & 2.32\% & 1.41\% \\
IRSE50     & 0.241 & 0.271 & 0.311 & 3.67\% & 1.82\% \\
FaceNet    & 0.409 & 0.467 & 0.483 & 2.38\% & 2.04\% \\
MobileFace & 0.302 & 0.288 & 0.402 & 5.36\% & 0.79\% \\
\bottomrule
\end{tabular}}

\end{table}

A threshold is the quantile of an impostor similarity distribution, and that
distribution depends on what the model is shown. We therefore calibrate
thresholds per front end on $500$ clean FFHQ images disjoint from
the evaluation set, taking the $(1-\mathrm{FAR})$ quantile over all impostor
pairs. The sample size is the $500$ images, not the $124{,}750$ pairs they
form, which are not independent observations. On aligned crops the inherited constants are close to our calibration
and produce a false-accept rate of $0.79$--$2.04\%$ against a nominal $1\%$; on
whole images they produce $2.32$--$5.36\%$ (Tab.~\ref{tab:thr}). Whole-image
embeddings of different people are more similar, sharing background and
framing, so the impostor distribution shifts up and a calibrated threshold must
rise with it (MobileFace: $0.402$ against $0.288$). The table the field
inherits is, in effect, an aligned-pipeline table on a whole-image pipeline.
FFHQ has no identity labels, so we repeat the calibration on CelebA-HQ with
one photograph per labelled identity, every pair a true impostor: the constants
admit $1.5$--$8.9\%$ on whole images against $0.5$--$2.0\%$ on aligned crops,
disjoint on three of four models (supplement).

\paragraph{The bias is opposite-signed across tasks}
Dodging counts similarity \emph{below} the threshold and impersonation
\emph{above}. A threshold too low for the pipeline therefore makes
$P_{\mathrm{dir}}$ look worse than it is for dodging and better for
impersonation. The supplement tests this sharp prediction on six independent
rows: correcting the calibration moves $K_1$ down in all four dodging rows and
up in both impersonation rows. Published dodging results therefore
\emph{understate} the protocol drop, ours roughly doubling under correct
calibration, while published impersonation results \emph{overstate} absolute
protection: our runs of two released methods fall from $30.2$ to $14.1$ and
from $31.6$ to $14.7$ once the threshold matches the pipeline (AMT-GAN, DiffAM;
Sec.~\ref{sec:published}). Both are our reproductions rescored, not published
numbers corrected; ours of DiffAM stands at $31.6$ against the $75.79$ its
paper reports, a gap we delimit rather than close (supplement). The two tasks, routinely tabulated side by side, are
biased in opposite directions.

\section{Two Loss Mechanisms}
\label{sec:mechanism}

The drop in Tab.~\ref{tab:k1} is not one phenomenon. Three controlled tests,
each changing exactly one thing, separate it into two.

\begin{figure}[t]
  \centering
  \includegraphics[width=\linewidth]{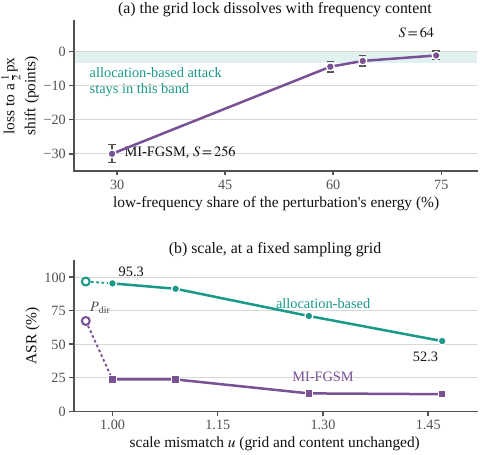}
  \caption{\textbf{The two loss mechanisms, side by side.}
  \textbf{(a)}~The grid lock belongs to the perturbation, not to the method:
  re-parameterising $\boldsymbol{\delta}$ at $S{\times}S$ and upsampling leaves
  MI-FGSM otherwise unchanged, and its loss to a half-output-pixel shift falls
  from $-30.0$ to $-1.2\pp$ as its low-frequency share rises, into the band the
  allocation-based attack occupies at every target.
  \textbf{(b)}~The other mechanism, with the sampling grid held fixed and the
  landmarks taken from the clean image so that detector drift cannot
  contribute (IR152). The allocation-based attack loses nothing at $u{=}1.00$
  and then falls monotonically with scale, which is a scale mechanism; MI-FGSM
  has already lost everything at $u{=}1.00$ (dotted, from its
  $P_{\mathrm{dir}}$ level) and is flat thereafter, which is a grid one.
  Whiskers are bootstrap $95\%$ CIs; thresholds calibrated per pipeline. Full
  sweeps, per target, in the supplement.}
  \label{fig:mechanism}
\end{figure}

\paragraph{The sampling grid, at fixed scale}
Resizing $256\to112$ samples the image at $112$ positions. Keeping scale, crop
and content fixed, we move only those: half an output pixel of translation
(about $1.14$ input pixels) and $1^\circ$ of rotation ($N{=}200$, thresholds
per pipeline). Under the half-pixel shift MI-FGSM loses $55.5\pp$ on
IR152 \ci{-62.5}{-48.5}, $44.5$--$52.0$ on IRSE50 and MobileFace and $11.0$ on
FaceNet; rotation costs it $31.5$--$36.0$ on the first three. The
allocation-based attack loses at most $3.5$ on any target under either. A
protection that dies from moving the sampling positions lives not in the image
content but in the aliasing phase of one resampling grid (supplement), of which
clean images show a milder form
\cite{azulay2019generalize,zhang2019antialias}.

\paragraph{The scale, at fixed grid}
We then render aligned crops at a controlled scale mismatch $u$, with landmarks
taken from the clean image so that detector drift cannot contribute
($N{=}300$). The allocation-based attack loses nothing at $u{=}1.00$ and then
falls monotonically as $u$ grows (Fig.~\ref{fig:mechanism}b), which identifies
scale as its mechanism and confirms Sec.~\ref{sec:decomp}. MI-FGSM has already lost
everything at $u{=}1.00$ and barely moves thereafter, so its $-26.6\pp$
resampling term is a grid effect, not a scale one.

\paragraph{Grid leakage}
FaceNet takes $160{\times}160$ inputs where the three surrogates take
$112{\times}112$, so under $P_{\mathrm{dir}}$ it is sampled on a different
grid; forcing it onto theirs moves MI-FGSM's $K_1$ from $-9.5$ to $-28.5$
\ci{-35.0}{-22.0} and leaves the allocation-based attack unchanged
(supplement). Part of what reads as cross-architecture transfer in direct
evaluation is surrogate and target sharing a resampling grid.

\paragraph{What distinguishes the regimes}
The attacks differ in frequency content: under a Gaussian split at
$\sigma{=}1.5$, $72\%$ of MI-FGSM's energy is high-frequency against $70\%$
low-frequency for the allocation-based attack. Re-running MI-FGSM with
$\boldsymbol{\delta}$ parameterised at a reduced resolution $S{\times}S$ and
upsampled, everything else fixed, sweeps that split at a fixed $\ell_\infty$
bound. The grid lock dissolves monotonically with it and $K_1$ converges on the
same place, from $-18.6$ at $S{=}256$ to about $-7.5$ for every low-frequency
variant against $-6.9$ for the allocation-based attack
(Fig.~\ref{fig:mechanism}a; sweep in the supplement). The regime is a property
of \emph{what the perturbation is made of}, not of the method that made it.
Those variants also inherit the scale mechanism, their scale sweep staying
monotone, and low frequency makes MI-FGSM \emph{transfer better} to FaceNet
under $P_{\mathrm{dir}}$ ($47.0\to70.7$ at $S{=}128$), grid leakage from the
other side.

\begin{table}[tb]
\caption{\textbf{Why does a budget-allocating attack survive: because its
budget sits on the face, or because it is low-frequency?} One optimiser, four
cells, the same \emph{mean} budget ($16/255$) throughout (FFHQ, $N{=}300$, four
leave-one-out targets, thresholds calibrated per pipeline). ``face-weighted''
is a Gaussian budget map centred on the landmark centroid, rescaled so that its
mean matches the uniform cell; $S$ is the resolution at which
$\boldsymbol{\delta}$ is parameterised before bilinear upsampling. The two
factors act on \emph{different} quantities. Concentrating the budget on the
face is what shrinks the protocol drop: $-21.7\rightarrow-6.9$ at $S{=}256$ and
$-9.8\rightarrow-4.2$ at $S{=}64$, and the face-weighted high-frequency cell
lands on the $-6.9$ we measure for the allocation-based attack itself. Low
frequency is what removes the grid lock: the half-pixel loss goes
$-31.2\rightarrow-0.4$ with the budget spread over the frame. The
allocation-based attack combines both. Caveat: at equal mean budget the
face-weighted cells reach a peak of $32/255$ against the uniform cells'
$16/255$, so the comparison equalises total budget, not peak amplitude.}
\label{tab:2x2}
\centering\small
\setlength{\tabcolsep}{4pt}
\resizebox{\linewidth}{!}{%
\begin{tabular}{@{}llrrrr@{}}
\toprule
Budget & $\boldsymbol{\delta}$ & $P_{\mathrm{dir}}$ & $P_{\mathrm{ali}}$ & $K_1$ & $\tfrac12$px \\
\midrule
uniform       & $S{=}256$ & 60.1 & 38.4 & $-21.7$ \ci{-24.5}{-18.9} & $-31.2$ \\
face-weighted & $S{=}256$ & 92.9 & 86.0 & $-6.9$ \ci{-8.8}{-5.1}    & $-16.6$ \\
uniform       & $S{=}64$  & 41.3 & 31.5 & $-9.8$ \ci{-12.2}{-7.5}   & $-0.4$ \\
face-weighted & $S{=}64$  & 91.8 & 87.6 & $-4.2$ \ci{-6.1}{-2.5}    & $-0.2$ \\
\bottomrule
\end{tabular}}

\end{table}

\paragraph{Two factors, not one}
Frequency content decides which regime an attack is in, not how much it loses.
A $2{\times}2$ design separates the two (one optimiser, one budget cap,
\{uniform, face-weighted\} $\times$ \{$S{=}256$, $S{=}64$\};
Tab.~\ref{tab:2x2}):
concentrating the budget on the face shrinks the drop, low frequency removes
the grid lock, and the allocation-based attack does both, which is why it
survives. TI-FGSM \cite{dong2019ti}, $91\%$ low-frequency, is the predicted
intermediate case: shift-robust ($-3.2\pp$) yet still losing margin to scale
(Tab.~\ref{tab:eps}).

\section{Pipeline-Aware Optimisation}
\label{sec:method}

Both regimes of Sec.~\ref{sec:mechanism} have the same origin: the evaluation
reuses the resampling operator the protection was optimised against. Both
therefore admit the same remedy, which is to put the deployed operator into the
objective. The three components below are standard, and optimising through a
differentiable alignment operator is itself not new
\cite{cherepanova2021lowkey}. The contribution is the diagnosis that says where
to apply them.

\paragraph{The three components}
Given landmarks $\ell$ we fit the similarity transform $M(\ell)$ to the ArcFace
template by the Umeyama estimator \cite{umeyama1991} and realise the warp with
\texttt{affine\_grid}/\texttt{grid\_sample}, so gradients flow from the loss
through the crop back to the full-resolution perturbation: the surrogate
embedding becomes $f(\mathcal{A}_{\ell}(x+\delta))$ rather than
$f(\mathrm{resize}(x+\delta))$. Only the warp is differentiated. The detector
is not, so we sample $\ell' = \ell + \mathcal{N}(0,\sigma^2)$ with
$\sigma = 1.5\,\mathrm{px}$ at every iteration \cite{athalye2018eot}; MTCNN and
RetinaFace disagree by $3.24\,\mathrm{px}$ on our data, more than both this
jitter and the displacement the protection induces. And because a protected
photograph may be consumed by either front end, we draw the aligned path with
probability $p = 0.5$ per iteration rather than optimising for it alone;
Tab.~\ref{tab:k1} confirms the protocol-pipeline numbers do not degrade. The
attack algorithm itself is unmodified: we rebind the name through which it
obtains embeddings, the same four-line change in the allocation-based attack
and in plain MI-FGSM.

\section{Experiments}
\label{sec:exp}

\subsection{Setup}
Four recognition models in leave-one-out black-box configuration (IR152,
IRSE50, FaceNet, MobileFace), three datasets (FFHQ, CelebA-HQ, LADN), two
detectors (MTCNN for optimisation and evaluation, RetinaFace for evaluation
only). FFHQ main results use $N{=}1000$; CelebA-HQ, LADN and the older
ablations $N{=}150$; the decomposition and further attacks $N{=}300$, the grid
tests $N{=}200$. The allocation-based attack \cite{parlayici2026ssmattack}
predicts a per-pixel budget with a state-space network and searches for a
perturbation respecting it, while plain MI-FGSM \cite{dong2018mifgsm} spends a
uniform $\epsilon = 16/255$ over the whole image ($50$ iterations). Our
modification touches only the operator computing the surrogate embedding
(Sec.~\ref{sec:method}).

\subsection{Main result}

Tab.~\ref{tab:k1} and the ablation of the supplement give the main result. On
these four recognizers the loss is removed, not merely reduced: under the
deployed pipeline the
allocation-based attack rises from $86.0$ to $98.2$, above its own
protocol-pipeline number, and MI-FGSM from $24.4$ to $74.6$. Two points deserve
emphasis.

\textbf{Alignment knowledge is doing real work.} Randomising scale alone, by an
EOT over rescaling with no landmarks, recovers $+6.15\pp$ of the $+12.22\pp$;
the actual alignment supplies the other $+6.08\pp$. The mechanism is not
reducible to generic input diversity, and the closest prior remedy
\cite{croft2025detectors} is of the first kind.

\textbf{The effect is base-attack agnostic.} The largest correction is on plain
MI-FGSM ($+50.2\pp$), which we did not design and do not defend, and it does
not trade one pipeline for the other: the pipeline-aware variant also improves
on $P_{\mathrm{dir}}$ ($+3.0\pp$), leaving $+9.2\pp$ as the alignment-specific
component.

\begin{table}[tb]
\caption{\textbf{Against a gallery of independent photographs.} CelebA-HQ,
$142$ subjects with $1$--$9$ other photographs each. \emph{own}: the subject's
own photograph; \emph{gallery}: success only if \emph{no} other photograph
matches. Thresholds per pipeline, rule and gallery size (FAR $1\%$ unless
marked) on $500$ identity-disjoint subjects. Margin shift is $\Delta m$ of
Sec.~\ref{sec:dcos}, \emph{before} the clean baseline is removed; subtracting
it by Eq.~\ref{eq:dcos} gives $\Delta\!\cos = -0.111$, $-0.018$, $+0.073$ and
$+0.128$ down the \emph{own} column, which reproduces the independently
measured FFHQ values of Tab.~\ref{tab:modern} to within $0.005$.
$^\ast$Direct success $100.0$: a lower bound.}
\label{tab:gallery}
\centering\small
\setlength{\tabcolsep}{3.4pt}
\resizebox{\columnwidth}{!}{%
\begin{tabular}{@{}lccc|cc@{}}
\toprule
 & \multicolumn{3}{c|}{$K_1$ (pp)} & \multicolumn{2}{c}{margin shift} \\
Protection & own & gallery & gallery$^{10\%}$ & own & gallery \\
\midrule
MI-FGSM         & \drop{$-37.5$} & \drop{$-27.5$} & \drop{$-34.3$} & $-0.154$ & $-0.135$ \\
\;$+$ \gd{ours} & $+1.0$  & $-1.2$  & $-2.1$  & $+0.030$ & $-0.011$ \\
Allocation      & \drop{$-8.0$}  & $-4.0^\ast$ & \drop{$-11.1$} & $-0.061$ & $-0.067$ \\
\;$+$ \gd{ours} & $+0.7$  & $-0.2$  & $-0.2$  & $+0.085$ & $+0.028$ \\
\bottomrule
\end{tabular}}

\end{table}

\textbf{A gallery of independent photographs.} We score dodging, as the
literature does, against the subject's own photograph, but a deployed system
matches against other photographs of the person
\cite{cherepanova2021lowkey,croft2025detectors}. Against every other
photograph of each subject (Tab.~\ref{tab:gallery}) the drop stays: MI-FGSM's
$K_1$ is $-27.5$, and the allocation-based attack's margin loss is unchanged.
Its smaller binary drop is a ceiling, not a recovery, and widens to $-11.1$ at
FAR $10\%$. The remedy still largely removes the loss.

\paragraph{An unseen detector}
\label{sec:unseen}
Optimisation uses MTCNN landmarks only; evaluated with RetinaFace the
improvement is $+13.60$ against $+12.22\pp$ (allocation) and $+47.8$ against
$+50.2$ (MI-FGSM). Five-point alignment maps any landmark estimate onto the
same template, absorbing the detector, which is why this does not contradict
the inter-detector variation Croft \etal\ \cite{croft2025detectors} report on
unaligned crops. JPEG sweeps and four image-quality measures (supplement) rule
out compression fragility and a gain bought with distortion.

\subsection{Published methods}
\label{sec:published}

We repeat the measurement on two published, released methods in their own
conventions and on their own task (impersonation): AMT-GAN \cite{hu2022amtgan}
and DiffAM \cite{sun2024diffam}, the latter over its four targets with the
hyperparameters its README sets (supplement). Both are
\emph{grid-robust}: the half-output-pixel shift costs them $2.0$ and $1.2\pp$,
placing them in the scale regime alongside the allocation-based attack, so the
regime does not track the method family. For impersonation the protocol drop
turns out to be a \emph{threshold artefact}. At the inherited thresholds both
appear to lose under the realistic pipeline, by $12.8$ and $5.7\pp$; at
calibrated thresholds AMT-GAN's $K_1$ is $-1.3$ \ci{-4.2}{+1.5}, not separable
from zero, and DiffAM's is $+8.1$, almost all of it from MobileFace, the model
whose two calibrated thresholds differ most (Tab.~\ref{tab:thr}): the scope
stated in Sec.~\ref{sec:intro}, these two carrying the threshold defect and not
the input defect. Absolute levels do change: AMT-GAN's reported success falls
from $30.2$ to $14.1$ once the threshold matches the pipeline, with no
reference to alignment at all.

\textbf{The one method that aligned.} TIP-IM \cite{yang2021tipim} is the only
method in Tab.~\ref{tab:survey} whose evaluation detects and aligns; its attack
is optimised on the aligned $112$ crop and pasted back through the inverse
transform. Our mechanism therefore makes a signed prediction, registered before
the run: a perturbation is locked to the grid it was optimised on, so TIP-IM
should lose under the \emph{whole-image} protocol, not under the deployed one.
It does. On the same photographs, thresholds and statistics as the rest of the
paper ($N{=}300$ per target, $\epsilon = 12/255$ as its code sets), TIP-IM
scores $17.8$ under $P_{\mathrm{dir}}$ and $42.9$ under $P_{\mathrm{ali}}$, so
$K_1 = \gain{+25.2\pp}$ \ci{+22.6}{+27.8}, positive on every target ($+13.6$ to
$+40.6$, each $p < 10^{-7}$). The protocol does not merely inflate: it
\emph{misranks}, deflating the one method that built the realistic front end;
the half-pixel shift, applied on the aligned side, confirms the grid lock
(supplement).

\textbf{The ranking changes, not only the numbers.} In a matched $N{=}300$ run
the protocol puts TIP-IM last of four by $42.4\pp$, while $P_{\mathrm{ali}}$
moves it inside the range of the other three; its error ($25.2\pp$) exceeds
the spread separating them ($14.9\pp$), so the protocol cannot order it.

\textbf{At a matched budget the sign of the gap differs across the family.}
Re-running TIP-IM at $\epsilon{=}16/255$ and $50$ iterations, the setting of
the three gradient baselines, gives $\Delta\!\cos = \gain{+0.184}$
\ci{+0.175}{+0.192} against $-0.046$ for MI-FGSM, $-0.029$ for TI-FGSM and
$+0.006$ for DI-FGSM (FFHQ, $N{=}300$, four targets, clean-landmark convention,
the only one available for TIP-IM; supplement). Equalising the budget therefore
does not remove the reversal, it isolates it: within one attack family at one
operating point, the protocol's error runs from $-0.05$ to $+0.18$. Budget and
iterations are matched here, computation is not, since TIP-IM evaluates ten
candidate targets per iteration.

\subsection{Beyond our four recognition models}
\label{sec:modern}

\begin{table}[tb]
\caption{\textbf{Three recognizers no attack used as a surrogate}: ArcFace
ResNet-50 and MobileFaceNet on WebFace600K \cite{zhu2021webface260m},
ResNet-100 on Glint360K \cite{an2022partialfc} (InsightFace
\texttt{buffalo\_l}, \texttt{buffalo\_s}, \texttt{antelopev2}). Identical
files, FFHQ, $N{=}1000$ per generation target, four leave-one-out generation
targets; the three recognizers above do the scoring and none of them was a
surrogate. Thresholds per front end and model; $\Delta\!\cos$ needs none (Eq.~\ref{eq:dcos}). The defect is
$2.2\times$ larger here than on the four standard recognizers for MI-FGSM
($-0.239$ pooled against $-0.109$) and $10.7\times$ larger for the
allocation-based attack ($-0.142$ against $-0.013$), so it is no artefact of
models reused since $2019$. The lower block gives $\Delta\!\cos$ only: its
success rates would be read against a different threshold cache than
Tab.~\ref{tab:k1}'s and differ from it by $1$ to $3\pp$ (supplement), where
$\Delta\!\cos$ is threshold-free and so comparable across both. $^{\dagger}$Floor. $^{\ddagger}$Near the ceiling.
The remedy does not transfer uniformly: it removes the gap on R50, overshoots
on MobileFaceNet and leaves $-0.091$ on R100/Glint360K, so the pooled $-0.031$
is partly cancellation.}
\label{tab:modern}
\centering\small
\setlength{\tabcolsep}{2.6pt}
\resizebox{\columnwidth}{!}{%
\begin{tabular}{@{}lccc|cc@{}}
\toprule
& \multicolumn{3}{c|}{$\Delta\!\cos$ per recognizer}
& \multicolumn{2}{c}{success rate} \\
\cmidrule(lr){2-4}\cmidrule(lr){5-6}
Protection & R50/WF & MBF/WF & R100/Glint
           & $P_{\mathrm{dir}}{\to}P_{\mathrm{ali}}$ & $K_1$ \\
\midrule
MI-FGSM         & \drop{$-0.250$} & \drop{$-0.211$} & \drop{$-0.256$}
                & $40.9 \to 1.9^{\dagger}$   & \drop{$-38.9$} \\
\;$+$ \gd{ours} & $-0.024$ & $+0.027$ & $-0.030$
                & $56.8 \to 47.3$   & \drop{$-9.5$} \\
Allocation      & \drop{$-0.153$} & \drop{$-0.098$} & \drop{$-0.176$}
                & $75.5 \to 35.1$   & \drop{$-40.4$} \\
\;$+$ \gd{ours} & $-0.052$ & $+0.050$ & \drop{$-0.091$}
                & $93.0 \to 83.9^{\ddagger}$ & \drop{$-9.1$} \\
\midrule
\multicolumn{6}{@{}l}{\emph{four standard recognizers, identical files}, $\Delta\!\cos$ \emph{only}}\\
MI-FGSM         & \multicolumn{3}{c|}{pooled \drop{$-0.109$}} & \multicolumn{2}{c}{Tab.~\ref{tab:k1} has the rates} \\
\;$+$ \gd{ours} & \multicolumn{3}{c|}{\gain{$+0.076$}}        & \multicolumn{2}{c}{} \\
Allocation      & \multicolumn{3}{c|}{\drop{$-0.013$}}        & \multicolumn{2}{c}{} \\
\;$+$ \gd{ours} & \multicolumn{3}{c|}{\gain{$+0.125$}}        & \multicolumn{2}{c}{} \\
\bottomrule
\end{tabular}}

\end{table}

The models compared so far date from $2016$--$2019$ and every attack used them
as surrogates. We therefore repeat the measurement outside that set: on three
modern recognizers no attack ever saw (Tab.~\ref{tab:modern}), and on two
commercial endpoints we neither built nor can inspect.

\textbf{The protocol defect grows on modern models.} On all three, plain
MI-FGSM's measurement all but collapses under a deployed front end, losing $90$
to $98\%$ of its $P_{\mathrm{dir}}$ protection against $56\%$ on the four
standard models, and the allocation-based attack loses $20$ to $69\%$ against
$7\%$. The surrogate-preprocessing dependence of Sec.~\ref{sec:measure} is
immaterial here: both MI-FGSM sets lose over $90\%$ on two models and $71\%$ on
the third. The gap is no artefact of models reused since $2019$: on current
ones it is larger, and the three disagree on \emph{how} large.

\textbf{Pipeline-aware optimisation transfers.} The paired gain under the
deployed pipeline is $+38.0$, $+56.3$ and $+41.3\pp$ on MI-FGSM and $+55.8$,
$+24.4$ and $+57.9\pp$ on the allocation-based attack, significant on every
target of every model; on MI-FGSM no photograph in $4000$, on any model, has
the baseline succeed where ours fails. The residual $K_1$ ($+1.4$ to
$-13.5\pp$) tracks the surrogate set rather than the remedy: adding one modern
recognizer to the surrogates (R50/WebFace600K, both targets held out) shrinks it
from $-4.7$ to $+3.7\pp$ \ci{+0.3}{+7.3} and from $-10.3$ to $-3.3$
\ci{-6.3}{-0.3}. Aligned-only optimisation beats mixing there but loses
$35.3$ and $19.0\pp$ on the whole image (supplement).

\textbf{The dataset, not the recognizer, sets how much is lost}, on these
recognizers as on the four standard ones (Sec.~\ref{sec:when}); the gains hold
on CelebA-HQ and LADN as well (\gain{$+62.7$} to \gain{$+74.0\pp$} on MI-FGSM,
\gain{$+16.2$} to \gain{$+66.5$} on the allocation-based attack), with at most
one photograph of $600$ against us in any of the twelve dataset--model
comparisons (supplement).

\textbf{Two commercial endpoints reproduce both defects.} On Face++
\texttt{compare}, used by prior work
\cite{hu2022amtgan,shamshad2023clip2protect,sun2024diffam}, $300$ unprotected
impostor pairs give $1.7\%$ \ci{0.5}{3.8} at the threshold documented as
FAR$=10^{-3}$. In all four configurations the protocol reports more protection
than the endpoint delivers, by $1.11$ to $5.50\times$ (MI-FGSM: $72.0$ against
a commercial $13.1$, deployed pipeline $13.0$); Amazon Rekognition repeats it
($100.0$ against $18.0$, $72.0$ against $8.0$). On both, the deployed pipeline
predicts the commercial outcome better ($7.2$ vs.\ $31.7\pp$ mean deviation;
$23.8$ vs.\ $48.2$) and the paired gains carry over (Face++ \gain{$+27.3$},
\gain{$+43.4\pp$}, $p < 10^{-7}$). Rekognition admits no single $1\%$-FAR
threshold, so we report the admissible range \cite{clopper1934}. Either
endpoint's deviation from a lab pipeline confounds the front end with transfer
to an unseen recognizer; the paired comparison does not, and is what we rest on.

\section{Conclusion and Limitations}

Published facial-privacy protections are measured on an input deployed
recognition systems never construct, at thresholds calibrated for a pipeline
the benchmark does not run, and in a unit that saturates at both ends of its
range. Correcting any of the three requires no new protected image. The input
defect is a resampling mismatch, not perturbation lost outside the crop, and
optimising through a differentiable alignment operator removes most of it
across five attacks, three datasets, seven recognition models, an unseen detector and two
commercial endpoints.

\paragraph{What an evaluation should fix} All three corrections cost nothing.
\emph{(i)~Normalise the face geometry}: the template fixes the inter-ocular
distance, where a whole-image resize leaves it to the dataset's framing, so
such a benchmark is not comparable across datasets (Sec.~\ref{sec:when}).
\emph{(ii)~Calibrate on the front end you evaluate with}, on identity-labelled
impostor pairs, print the threshold, and tabulate dodging and impersonation
separately. \emph{(iii)~Report a measure that does not saturate}: the paired
cosine shift of Eq.~\ref{eq:dcos} needs no threshold, so a comparison reported
in it survives a change of calibration without being re-run. Our code (\repo) makes the
calibration, not a constant, what is reused.

\paragraph{Limitations}
The scope is narrower than the title. \emph{For impersonation there is no
protocol drop at calibrated thresholds}: AMT-GAN's $K_1$ is indistinguishable
from zero and DiffAM's positive, and our DiffAM runs stay below the published
level ($31.6$ against $75.79$), a gap its five reference makeups close only
$15\%$ of. \emph{One recognizer disagrees}: on MobileFace
$\Delta\!\cos$ has the opposite sign on five of six dataset and attack
combinations. Its two calibrations diverge far more than the others'
($-0.102$ against $-0.016$ to $-0.033$), and standardising by the impostor
distribution removes half of the anomaly on the undefended attacks but little
of it on the pipeline-aware variants (supplement); we report pooled values
with and without it. \emph{The allocation-based attack's
own drop is small} ($6.9\pp$, $2.1$ in the field convention), partly a ceiling
effect, so the claim rests on MI-FGSM, the threshold correction and the
pipeline-aware gain; MI-FGSM's drop depends on its surrogates' preprocessing
($56\%$ against $36\%$). \emph{The matched-budget comparison equalises
$\ell_\infty$ and iterations but not computation}: TIP-IM evaluates ten
candidate targets per iteration where the gradient baselines evaluate one.
\emph{Against a gallery of independent photographs} the drop survives, but
under the mean rather than the maximum rule the remedy leaves $-8.3\pp$ on
MI-FGSM. \emph{On modern recognizers the remedy does not close the gap
uniformly}, overshooting on one and leaving $-0.091$ on another
(Tab.~\ref{tab:modern}). Six further caveats are in the supplement.

\section*{Acknowledgment}

A large language model (Anthropic Claude) and a grammar checker (Grammarly)
were used for copy-editing the manuscript and while writing the measurement
and plotting scripts, which the authors then ran and inspected. Every claim,
number and citation was written or verified by the authors against the run or
the source it refers to; no number in this paper was produced by a language
model. The research question, the experimental design and the conclusions are
the authors' own.

\bibliographystyle{IEEEtran}
\bibliography{refs}

\begin{thebibliography}{10}

\bibitem{shan2020fawkes}
S.~Shan, E.~Wenger, J.~Zhang, H.~Li, H.~Zheng, and B.~Y. Zhao, ``{Fawkes}:
  Protecting privacy against unauthorized deep learning models,'' in {\em
  USENIX Security Symposium}, pp.~1589--1604, 2020.

\bibitem{cherepanova2021lowkey}
V.~Cherepanova, M.~Goldblum, H.~Foley, S.~Duan, J.~P. Dickerson, G.~Taylor, and
  T.~Goldstein, ``{LowKey}: Leveraging adversarial attacks to protect social
  media users from facial recognition,'' in {\em ICLR}, 2021.

\bibitem{yang2021tipim}
X.~Yang, Y.~Dong, T.~Pang, H.~Su, J.~Zhu, Y.~Chen, and H.~Xue, ``Towards face
  encryption by generating adversarial identity masks,'' in {\em ICCV},
  pp.~3897--3907, 2021.

\bibitem{hu2022amtgan}
S.~Hu, X.~Liu, Y.~Zhang, M.~Li, L.~Y. Zhang, H.~Jin, and L.~Wu, ``Protecting
  facial privacy: Generating adversarial identity masks via style-robust makeup
  transfer,'' in {\em CVPR}, pp.~15014--15023, 2022.

\bibitem{sun2024diffam}
Y.~Sun, L.~Yu, H.~Xie, J.~Li, and Y.~Zhang, ``{DiffAM}: Diffusion-based
  adversarial makeup transfer for facial privacy protection,'' in {\em CVPR},
  pp.~24584--24594, 2024.

\bibitem{deng2019arcface}
J.~Deng, J.~Guo, N.~Xue, and S.~Zafeiriou, ``{ArcFace}: Additive angular margin
  loss for deep face recognition,'' in {\em CVPR}, pp.~4690--4699, 2019.

\bibitem{zhang2016mtcnn}
K.~Zhang, Z.~Zhang, Z.~Li, and Y.~Qiao, ``Joint face detection and alignment
  using multitask cascaded convolutional networks,'' {\em IEEE Signal
  Processing Letters}, vol.~23, no.~10, pp.~1499--1503, 2016.

\bibitem{deng2020retinaface}
J.~Deng, J.~Guo, E.~Ververas, I.~Kotsia, and S.~Zafeiriou, ``{RetinaFace}:
  Single-shot multi-level face localisation in the wild,'' in {\em CVPR},
  pp.~5203--5212, 2020.

\bibitem{karras2019stylegan}
T.~Karras, S.~Laine, and T.~Aila, ``A style-based generator architecture for
  generative adversarial networks,'' in {\em CVPR}, pp.~4401--4410, 2019.

\bibitem{shamshad2023clip2protect}
F.~Shamshad, M.~Naseer, and K.~Nandakumar, ``{CLIP2Protect}: Protecting facial
  privacy using text-guided makeup via adversarial latent search,'' in {\em
  CVPR}, pp.~20595--20605, 2023.

\bibitem{advcpg2025}
J.~Wang, H.~Zhang, and Y.~Yuan, ``{Adv-CPG}: A customized portrait generation
  framework with facial adversarial attacks,'' in {\em CVPR}, pp.~21001--21010,
  2025.

\bibitem{liu2024advdiffusion}
D.~Liu, X.~Wang, C.~Peng, N.~Wang, R.~Hu, and X.~Gao, ``{Adv-Diffusion}:
  Imperceptible adversarial face identity attack via latent diffusion model,''
  {\em Proceedings of the AAAI Conference on Artificial Intelligence}, vol.~38,
  no.~4, pp.~3585--3593, 2024.

\bibitem{li2024gift}
M.~Li, J.~Wang, H.~Zhang, Z.~Zhou, S.~Hu, and X.~Pei, ``Transferable
  adversarial facial images for privacy protection,'' in {\em ACM MM}, 2024.

\bibitem{li2025tca}
W.~Li, Z.~Zhang, X.~Lan, and D.~Jiang, ``Transferable adversarial face attack
  with text controlled attribute,'' {\em Proceedings of the AAAI Conference on
  Artificial Intelligence}, vol.~39, no.~5, pp.~4977--4985, 2025.

\bibitem{salar2025weakening}
A.~Salar, Q.~Liu, Y.~Tian, and G.~Zhao, ``Enhancing facial privacy protection
  via weakening diffusion purification,'' in {\em CVPR}, pp.~8235--8244, 2025.

\bibitem{le2025diffprivate}
M.-h. Le and N.~Carlsson, ``{DiffPrivate}: Facial privacy protection with
  diffusion models,'' {\em Proceedings on Privacy Enhancing Technologies},
  vol.~2025, no.~2, pp.~54--70, 2025.

\bibitem{wang2026protego}
Z.~Wang, S.~Yang, J.~Lu, and K.-H. Chow, ``{Protego}: User-centric
  pose-invariant privacy protection against face recognition-induced digital
  footprint exposure,'' in {\em CVPR}, 2026.

\bibitem{athalye2018obfuscated}
A.~Athalye, N.~Carlini, and D.~Wagner, ``Obfuscated gradients give a false
  sense of security: Circumventing defenses to adversarial examples,'' in {\em
  ICML}, 2018.

\bibitem{tramer2020adaptive}
F.~Tram{\`e}r, N.~Carlini, W.~Brendel, and A.~Madry, ``On adaptive attacks to
  adversarial example defenses,'' in {\em NeurIPS}, 2020.

\bibitem{carlini2019evaluating}
N.~Carlini, A.~Athalye, N.~Papernot, W.~Brendel, J.~Rauber, D.~Tsipras,
  I.~Goodfellow, A.~Madry, and A.~Kurakin, ``On evaluating adversarial
  robustness.'' arXiv:1902.06705, 2019.

\bibitem{goodfellow2015fgsm}
I.~J. Goodfellow, J.~Shlens, and C.~Szegedy, ``Explaining and harnessing
  adversarial examples,'' in {\em ICLR}, 2015.

\bibitem{madry2018pgd}
A.~Madry, A.~Makelov, L.~Schmidt, D.~Tsipras, and A.~Vladu, ``Towards deep
  learning models resistant to adversarial attacks,'' in {\em ICLR}, 2018.

\bibitem{guo2018countering}
C.~Guo, M.~Rana, M.~Ciss{\'e}, and L.~van~der Maaten, ``Countering adversarial
  images using input transformations,'' in {\em ICLR}, 2018.

\bibitem{zhang2023lfap}
J.~Zhang, Q.~Yi, D.~Lu, and J.~Sang, ``Low-mid adversarial perturbation against
  unauthorized face recognition system,'' {\em arXiv:2206.09410}, 2023.

\bibitem{guo2024jpegrobust}
K.~Guo, F.~Zhou, H.~Ling, P.~Li, and H.~Liu, ``Improving the {JPEG}-resistance
  of adversarial attacks on face recognition by interpolation smoothing,'' {\em
  arXiv:2402.16586}, 2024.

\bibitem{xie2018mitigating}
C.~Xie, J.~Wang, Z.~Zhang, Z.~Ren, and A.~Yuille, ``Mitigating adversarial
  effects through randomization,'' in {\em ICLR}, 2018.

\bibitem{athalye2018eot}
A.~Athalye, L.~Engstrom, A.~Ilyas, and K.~Kwok, ``Synthesizing robust
  adversarial examples,'' in {\em ICML}, pp.~284--293, 2018.

\bibitem{azulay2019generalize}
A.~Azulay and Y.~Weiss, ``Why do deep convolutional networks generalize so
  poorly to small image transformations?,'' {\em Journal of Machine Learning
  Research}, vol.~20, no.~184, pp.~1--25, 2019.

\bibitem{zhang2019antialias}
R.~Zhang, ``Making convolutional networks shift-invariant again,'' in {\em
  ICML}, pp.~7324--7334, 2019.

\bibitem{croft2025detectors}
R.~Croft, B.~Du, D.~Joseph, and S.~Kumar, ``Investigating adversarial
  robustness against preprocessing used in blackbox face recognition,'' in {\em
  DICTA}, pp.~1--8, 2025.

\bibitem{zhou2024rethinking}
F.~Zhou, Q.~Zhou, B.~Yin, H.~Zheng, X.~Lu, L.~Ma, and H.~Ling, ``Rethinking
  impersonation and dodging attacks on face recognition systems,'' in {\em
  Proceedings of the 32nd ACM International Conference on Multimedia},
  pp.~2487--2496, 2024.

\bibitem{parlayici2026ssmattack}
T.~Parlay{\i}c{\i}, E.~Aptoula, and Y.~Sayg{\i}n, ``{SSM-Attack}: Learning
  where to perturb for black-box face privacy.'' arXiv:2610.17323, 2026.

\bibitem{wang2018cosface}
H.~Wang, Y.~Wang, Z.~Zhou, X.~Ji, D.~Gong, J.~Zhou, Z.~Li, and W.~Liu,
  ``{CosFace}: Large margin cosine loss for deep face recognition,'' in {\em
  CVPR}, 2018.

\bibitem{chen2018mobilefacenets}
S.~Chen, Y.~Liu, X.~Gao, and Z.~Han, ``{MobileFaceNets}: Efficient {CNNs} for
  accurate real-time face verification on mobile devices,'' in {\em Chinese
  Conference on Biometric Recognition (CCBR)}, 2018.

\bibitem{yin2021advmakeup}
B.~Yin, W.~Wang, T.~Yao, J.~Guo, Z.~Kong, S.~Ding, J.~Li, and C.~Liu,
  ``{Adv-Makeup}: A new imperceptible and transferable attack on face
  recognition,'' in {\em IJCAI}, pp.~1252--1258, 2021.

\bibitem{liu2026diffprotect}
J.~Liu, C.~P. Lau, Z.~Guo, Y.~Guo, Z.~Wang, and R.~Chellappa, ``{DiffProtect}:
  Generative adversarial examples using diffusion models for facial privacy
  protection,'' {\em Pattern Recognition}, vol.~173, p.~112780, 2026.

\bibitem{umeyama1991}
S.~Umeyama, ``Least-squares estimation of transformation parameters between two
  point patterns,'' {\em IEEE Transactions on Pattern Analysis and Machine
  Intelligence}, vol.~13, no.~4, pp.~376--380, 1991.

\bibitem{karras2018pggan}
T.~Karras, T.~Aila, S.~Laine, and J.~Lehtinen, ``Progressive growing of {GANs}
  for improved quality, stability, and variation,'' in {\em ICLR}, 2018.

\bibitem{gu2019ladn}
Q.~Gu, G.~Wang, M.~T. Chiu, Y.-W. Tai, and C.-K. Tang, ``{LADN}: Local
  adversarial disentangling network for facial makeup and de-makeup,'' in {\em
  ICCV}, pp.~10481--10490, 2019.

\bibitem{efron1979bootstrap}
B.~Efron, ``Bootstrap methods: another look at the jackknife,'' {\em The Annals
  of Statistics}, vol.~7, no.~1, 1979.

\bibitem{mcnemar1947}
Q.~McNemar, ``Note on the sampling error of the difference between correlated
  proportions or percentages,'' {\em Psychometrika}, vol.~12, pp.~153--157,
  1947.

\bibitem{dong2019ti}
Y.~Dong, T.~Pang, H.~Su, and J.~Zhu, ``Evading defenses to transferable
  adversarial examples by translation-invariant attacks,'' in {\em CVPR},
  pp.~4312--4321, 2019.

\bibitem{dong2018mifgsm}
Y.~Dong, F.~Liao, T.~Pang, H.~Su, J.~Zhu, X.~Hu, and J.~Li, ``Boosting
  adversarial attacks with momentum,'' in {\em CVPR}, pp.~9185--9193, 2018.

\bibitem{zhu2021webface260m}
Z.~Zhu, G.~Huang, J.~Deng, Y.~Ye, J.~Huang, X.~Chen, J.~Zhu, T.~Yang, J.~Lu,
  D.~Du, and J.~Zhou, ``{WebFace260M}: A benchmark unveiling the power of
  million-scale deep face recognition,'' in {\em CVPR}, 2021.

\bibitem{an2022partialfc}
X.~An, J.~Deng, J.~Guo, Z.~Feng, X.~Zhu, J.~Yang, and T.~Liu, ``Killing two
  birds with one stone: Efficient and robust training of face recognition
  {CNNs} by partial {FC},'' in {\em CVPR}, 2022.

\bibitem{clopper1934}
C.~J. Clopper and E.~S. Pearson, ``The use of confidence or fiducial limits
  illustrated in the case of the binomial,'' {\em Biometrika}, vol.~26, no.~4,
  pp.~404--413, 1934.

\end{thebibliography}

\clearpage
\includepdf[pages=-]{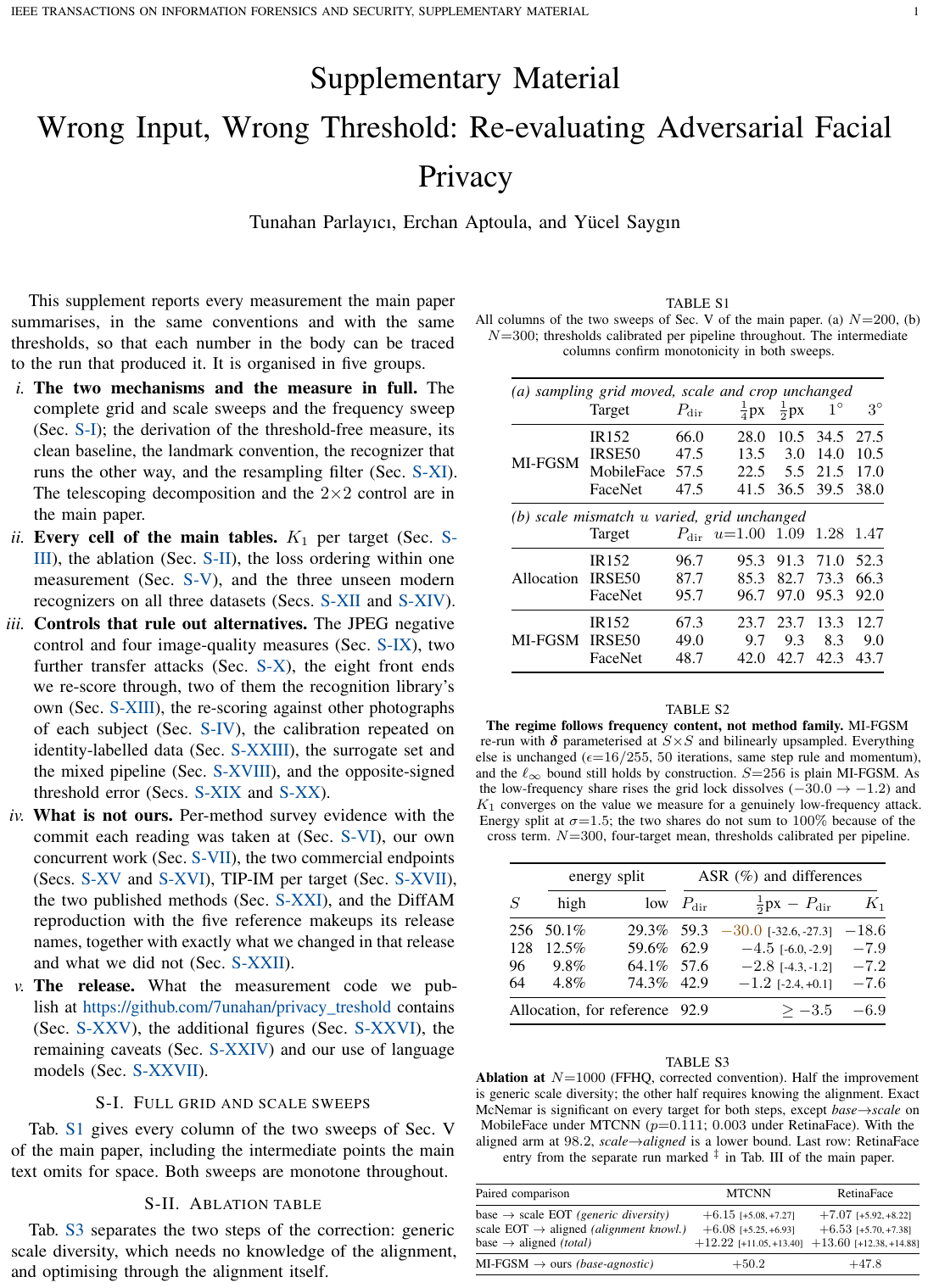}

\end{document}